\documentclass{bmvc2k}
\usepackage{bmvc2k_natbib}
\title{MALPOLON: A Framework for \\ Deep Species Distribution Modeling}

\addauthor{Theo Larcher}{larchertheo@gmail.com}{1}
\addauthor{Lukas Picek}{lukaspicek@gmail.com}{1,2}
\addauthor{Benjamin Deneu}{benjamin.deneu@wsl.ch}{3}
\addauthor{Titouan Lorieul}{titouan.lorieul@gmail.com}{}
\addauthor{Maximilien Servajean}{maximiliense@gmail.com}{1,4}
\addauthor{Alexis Joly}{alexis.joly@inria.fr}{1}
\addinstitution{
 INRIA\\
 \textit{Montpellier, France}\\
}

\addinstitution{
 University of West Bohemia\\
 \textit{Pilsen, Czechia}\\
}

\addinstitution{
Swiss Federal Institute for Forest, \\Snow and Landscape Research\\
\textit{Birmensdorf, Zurich, Switzerland}
}

\addinstitution{
 Université Paul Valéry, Montpellier\\
 \textit{Montpellier, France}\\
}

\runninghead{Larcher et al.}{MALPOLON}

\begin{document}

\maketitle

\begin{abstract}
    This paper describes a deep-SDM framework, MALPOLON. Written in Python and built upon the PyTorch library, this framework aims to facilitate training and inferences of deep species distribution models (deep-SDM) and sharing for users with only general Python language skills (e.g., modeling ecologists) who are interested in testing deep learning approaches to build new SDMs. More advanced users can also benefit from the framework's modularity to run more specific experiments by overriding existing classes while taking advantage of press-button examples to train neural networks on multiple classification tasks using custom or provided raw and pre-processed datasets.
    The framework is open-sourced on
    \href{https://github.com/plantnet/malpolon/tree/main}{GitHub} and \href{https://pypi.org/project/malpolon/}{PyPi} 
    along with extensive documentation and examples of use in various scenarios.
    MALPOLON offers straightforward installation, YAML-based configuration, parallel computing, multi-GPU utilization, baseline and foundational models for benchmarking, and extensive tutorials/documentation, aiming to enhance accessibility and performance scalability for ecologists and researchers.
    % exploring deep learning approaches in SDMs. 
\end{abstract}

\section{Introduction}
\label{sec:intro}
% Problem Definition
Species distribution models (SDMs) are popular numerical tools  \cite{araújo2019sdmassessment, srivastava2019sdminvasive} used for predicting the distribution of species over a geographic area and temporal frame, by trying to find correlations between environmental data and observation data. These models can be used to describe how environmental conditions or anthropogenic actions influence the occurrence or abundance of species and allow for future predictions  \cite{guisan2005sdmoverview}.
% How it is/was solved 
Historically, this work has been carried out by ecologists, botanists, or environmental researchers with a strong statistical background and expertise with mechanistic (\textit{process-based}) or statistical algorithms. Common methods used in species distribution modeling include
BIOMOD  \cite{thuiller2009biomod}, Maximum Entropy (MAXENT)  \cite{phillips2008modeling,phillips2004maxent,phillips2004maxent_ecomod}, 
Generalized Linear Models (GLM), 
Generalized additive model (GAM)  \cite{moisen2006gam_gboost},
Random forests  \cite{cutler2007rfclassif}, Boosted regression trees, Gradient Boosting, and Support Vector Machines. 
However, recent research has highlighted the potential of using deep learning methods such as CNNs to perform Presence Only (PO) or Presence Absence (PA) species prediction with competitive performances  \cite{deneu2021cnnsdm,deneu2018glc2018,botella2018deepsdm,chen2017dmse}.  
% Such models, called DeepSDMs, allow modeling the distribution of thousands of species in a single joint model, taking input from very high-resolution and complex data such as remote-sensing images or time series. 
So-called deep-SDM models offer an advantage in predicting suitability scores for many species using a unified model rather than relying on multiple individual single-species models. They excel in capturing relations among diverse environmental data  \cite{deneu2021cnnsdm,estopinan2022orchid, botella2023glc}.
\begin{figure}
    \centering
    {\hspace{4pt}}
    \includegraphics[width=0.925\linewidth, trim={62 40 50 30}, clip]{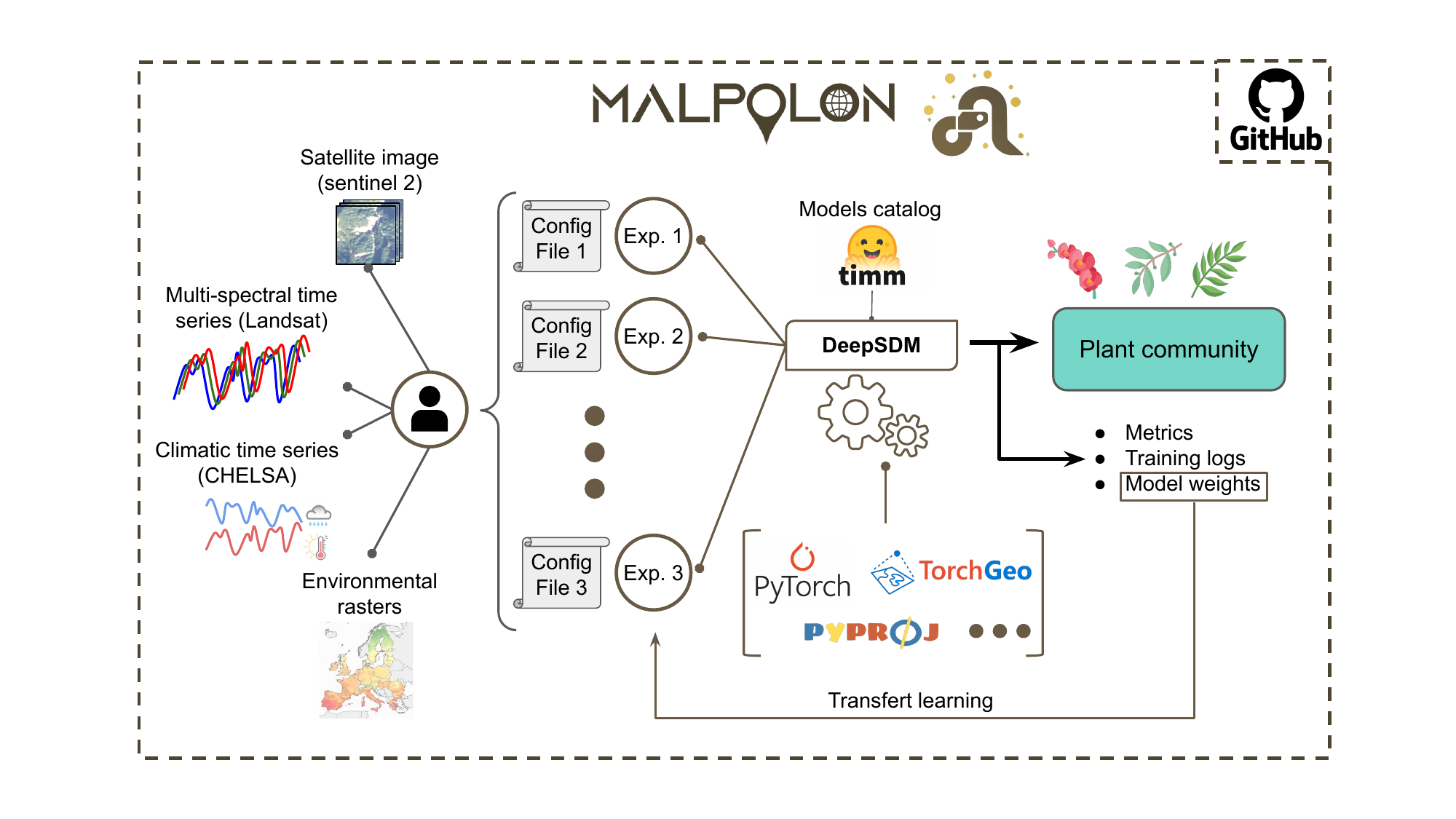}
    \caption{\textbf{Graphical abstract}. MALPOLON allows straightforward: (i) loading of various predictors, such as environmental rasters (e.g., land cover, human footprint), remote sensing data (e.g., Sentinel-2A and Landsat), and bioclimatic time-series, (ii) use of  geospatial foundational models (e.g., SatCLIP, GeoCLIP), (iii) model training with the press of a button.}
    \label{fig:figure_1}
    \vspace{-0.1cm}
\end{figure}
% What are the problems  ...
Although promising, such models are often developed and trained independently, lacking a shared foundation, which complicates reproducibility and accessibility. Furthermore, within the environmental research community focused on SDMs, the predominant tools and packages are rooted in the CPU-oriented language R. Many established SDM frameworks and widely-used statistical algorithms  \cite{adde2023nestedsdm, foster2024risdm, trevor2014bioclim} have historically been developed within R, fostering community growth and the evolution of new methodologies over time. However, this reliance on R presents limitations in terms of computational capabilities and access to emerging methods like Deep Neural Networks (DNNs), which are predominantly implemented in Python and C++.
% This is how we solve it ...

Despite being limited in number, recent development efforts have yielded a few solutions to establish and support deep learning SDM frameworks. For example,  \citet{pichler2021sjsdm} introduced \href{https://github.com/TheoreticalEcology/s-jSDM}{sjSDM}, which leverages PyTorch within R, enabling GPU utilization. However, integrating PyTorch code into an R package can decrease code and memory efficiency while also increasing the complexity and cognitive load associated with coding.
\citet{lu2017irl-imitation} proposed a Python implementation of Maxent through Deep Inverse Reinforcement Learning \cite{WulfmeierOP15}, however despite enabling a direct utilization of GPU capabilities via Tensorflow the framework relies on reward maps and does not handle observation / environmental predictor pairs as input data.
\citet{deepbiosphere} have developed a deep learning framework to perform plant prediction using citizen science and remote sensing data from California, enabling observation of species distribution shifts. However, this work is more focused on the model they propose and less about the framework in itself. While it is well-documented with comments, tutorials, and troubleshooting guidance, users need a solid understanding of Python and PyTorch to adapt it for custom use cases, as the framework is not modular.
% Our contributions ...

In light of that, we have developed MALPOLON, a new framework for deep species distribution modeling (see \hyperref[fig:figure_1]{Figure~\ref{fig:figure_1}}).
% As illustrated on Figure 1, 
It allows training various types of multi-modal SDMs based on deep learning using geolocated observations combined with a variety of input variables such as environmental rasters (e.g., land cover, human footprint, etc.), remote sensing images (e.g., Sentinel-2A  \cite{sentinel2a} and Landsat  \cite{landsat}), and time series (e.g., Landsat, and bioclimatic). MALPOLON is built on top of \href{https://lightning.ai/docs/pytorch/stable/}{PyTorch Lightning} and is distributed through  \href{https://pypi.org/project/malpolon/}{PyPi}  and available on \href{https://github.com/plantnet/malpolon/tree/main}{GitHub}, alongside installation guidelines, tutorials, and documentation.

% Praise more:
%   - One-liner execution
%   - Talk about the new experiments

% Although users do not need to know the inner mechanics of these frameworks as MALPOLON's experiments are fully parametrizable via YAML configuration files.

% \section{Related work}
% Existing framework to help researchers train SDM are mostly written in the CPU-oriented language R and offer a range of non-deep machine learning algorithms.

% \noindent \textbf{Nested-SDM}  \cite{adde2023nestedsdm} is an open-sourced R package which aims at addressing the issue of spatial niche truncation by proposing a pipeline capable of merging SDM models working with species and covariate data retrieved from global to regional scales.

% \noident \textbf{RISDM}  \cite{foster2024risdm}  is an open-sourced R package bringing  data integration methods to apply weighted joint likelihoods, correlation and covariates algorithms to build SDM with heterogeneous data.
% % 

% % Mostly written in R ...

% % Point out negatives ...

% % Methods ...

% \begin{figure}[t]
%     \centering
%     \includegraphics[width=\linewidth]{resources/figures/meta_classes_detailed}
%     \caption{\textbf{Meta-class diagrams of MALPOLON}. 
%     The \textit{Experiments} consist of different use cases with pre-written plug-and-play examples and include training, inference, etc.
%     The \textit{Engine} contains everything important for datasets and models loading and usage.
%     The \textit{Toolbox} provides a collection of useful scripts to perform data pre-processing.}
%     \label{fig:malpolon_meta_class}
% \end{figure}

\begin{figure}[t]
    \centering
    \includegraphics[width=\linewidth, trim={0 168 0 50}, clip]{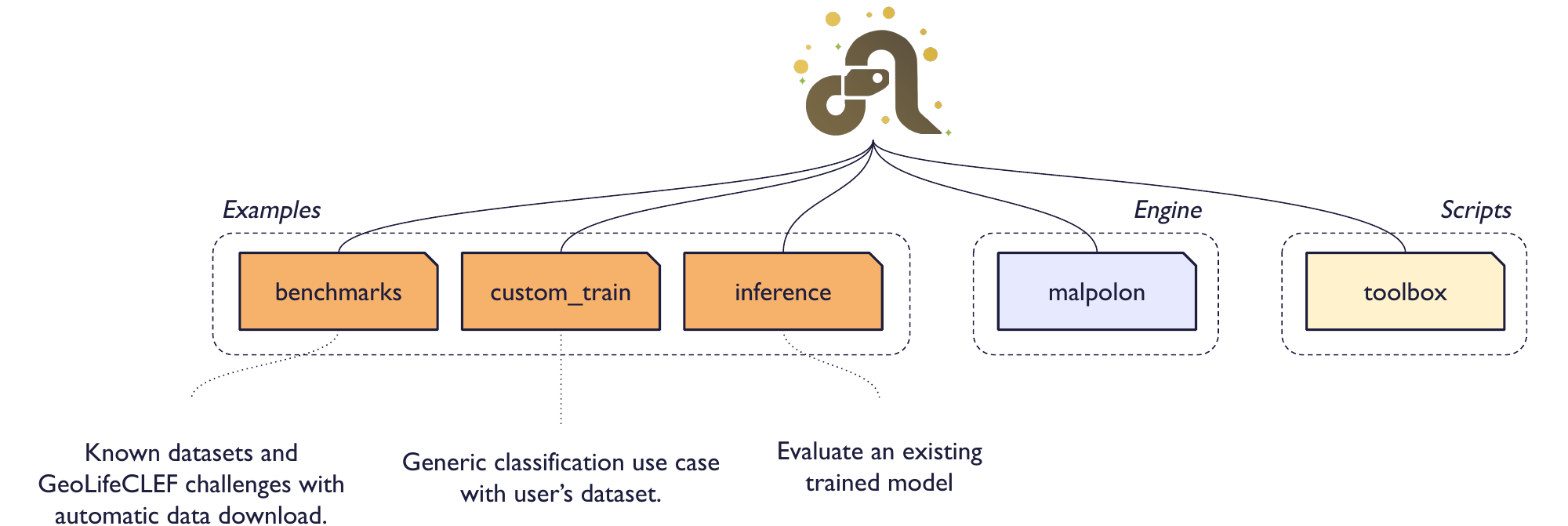}
    \caption{\textbf{Macro structure of MALPOLON}. 
    The \textit{Examples} consist of different use case experiments with pre-written plug-and-play examples and include training, inference, etc.
    The \textit{Engine} contains everything important for datasets and models loading and usage.
    The \textit{Toolbox} provides a collection of useful scripts to perform data pre-processing.}
    \label{fig:malpolon_meta_class}
        \vspace{-0.1cm}
\end{figure}

\section{The framework}
MALPOLON is a new Python-based framework for deep species distribution modeling, which is designed not only for ML researchers but also for people without extensive knowledge of Python and PyTorch.
The framework is built on top of PyTorch Lightning, which makes it highly modular and customizable, as its classes can be re-defined to better control the functions handling the datasets, data module, training loops, and optimizer. 
MALPOLON is compatible with \href{https://torchgeo.readthedocs.io/en/stable/}{TorchGeo}, a PyTorch domain library providing datasets, samplers, transforms, and pre-trained models specific to geospatial data. Furthermore, it allows using various types of neural networks ranging from simple MLP to more complex Transformer- and Convolutional-based architectures
% \footnote{We provide all models available in \href{https://pytorch.org/vision/stable/models.html}{Torchvision} and \href{https://huggingface.co/timm}{TIMM}}
. We also provide a variety of geospatial foundational models such as GeoCLIP~\cite{vivanco2024geoclip} and SatCLIP \cite{klemmer2023satclip}.
More importantly, MALPOLON offers straightforward access to various standards predictors as well as new ones, including but not limited to satellite data (e.g., Sentinel-2A and Landsat), climatic rasters (e.g., temperature and precipitation), and environmental rasters (e.g., soil grids, land cover, and human footprint). A collection of standalone data processing scripts is also provided under the name of \textit{toolbox}. We visualize MALPOLON's meta-class diagram in \hyperref[fig:malpolon_meta_class]{Figure~\ref{fig:malpolon_meta_class}}.

\subsection{Tools and Methods}

Experiments are framed around three entities: (i) \textbf{dataset}, (ii) \textbf{data module}, and (iii) \textbf{model}, that are later fed into a custom PyTorch Lightning pipeline.

The \textbf{datasets} define how data and labels are queried and returned when iterating through the dataset.
 The framework provides specialized dataset classes tailored for handling geospatial rasters, image patches, or combinations of both. Torchgeo-based datasets for geospatial rasters are particularly flexible with different Coordinate Reference Systems (CRS). They allow observations to be matched with rasters even if they are in different CRSs, eliminating the need for coordinate transformations or raster deformations.
% The \textbf{data module} is in charge of creating PyTorch \verb|dataloaders| for all or either training, validation, or test sets and allows data pre-processing and loading based on a given dataset. These PyTorch objects are then forward to a PyTorch Lightning \verb|Trainer|, in charge of handling the pipeline.

The \textbf{data module} is responsible for creating PyTorch \verb|dataloaders| for training, validation, and test sets, while also managing data pre-processing and loading according to the specified dataset. Once created, these PyTorch objects are then forwarded to a PyTorch Lightning \verb|Trainer|, which handles the pipeline.
% The \textbf{model} is in charge of holding the architecture of the model to train, re-defining the \verb|forward| and \verb|step| functions to match the dataset's tuple output to pass it on the model computation, and calling the model's metrics to compute and log them regularly.

The \textbf{model} holds the architecture to train, redefining the \verb|forward| and \verb|step| functions to match the dataset's tuple output for model computation. Additionally, it handles consistent computation and performance metrics logging throughout the optimization process.

By default, all experiments print a summary of the architecture, set the checkpoint saving strategy to validation loss decrease, and save the experiments' metrics and loss in CSV and Tensorboard files respectively. 
% \footnote{Any experiment logger callback can be added/removed in the main script.}. 
While training, a progress bar with the training status and training time estimation is provided. Furthermore, the model's loss and performance metrics are logged regularly to local \href{https://www.tensorflow.org/tensorboard?hl=fr}{Tensorboard} files, which allows visualizing the training by using the bash command \verb|tensorboard --logdir logs/|.
After training, the validation metrics of the best checkpoint are displayed, and the last output files are logged in the output directory and, if selected, "uploaded" to \href{https://www.tensorflow.org/tensorboard}{Tensorboard} or \href{https://wandb.ai/}{Weights \& Biases}.

\begin{figure}[t]
    \centering
    \includegraphics[width=\linewidth, trim={0 50 40 40}, clip]{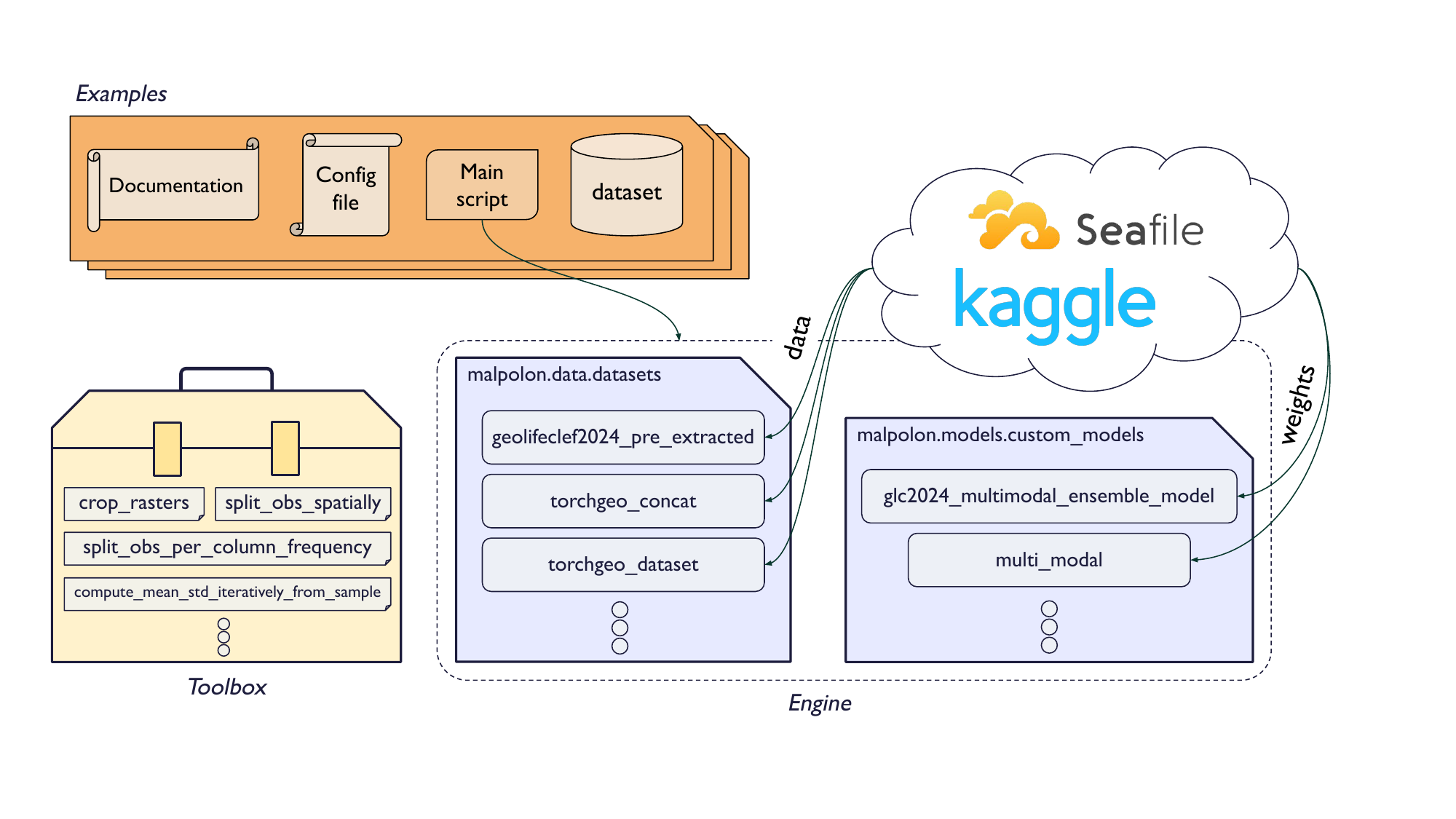}
    \caption{\textbf{Main components of MALPOLON}.
    The framework contains custom datasets and models (in \textit{blue}), which data and weights are automatically retrieved from remote servers. The toolbox (in \textit{yellow}), provides standalone data processing scripts. Examples (in \textit{orange}) are provided when cloning the GitHub project and interact with the engine to run models for training or inference.}
    \label{fig:main_components}
\end{figure}

\subsection{Data Availability}

Recent advances in species distribution modeling have shown that on top of environmental data, time-series and satellite imagery also contribute strongly to new model performances \cite{stewart2021climateXtremeTS,estopinan2022orchid,morand2023cnnsdmocean}.
However, the data is usually extremely disk space-demanding and time-consuming to pre-process, especially when combining multiple modalities and remote sensing data at the same time. Hence why, efforts have been made to synthesize and compress such data information in more conveniently compact data formats such as EcoDataCube \cite{ecodatacube}, OpenDataCune \cite{killough2018odc}, and \href{https://eurodatacube.com/}{EuroDataCube} which still can include gigabytes or terabytes of data.
In light of that, we build MALPOLON to provide straightforward access to all standard and even more complex data (see Figure \ref{fig:main_components}). So far, we allow loading the data from Sentinel-2A, Landsat, CHELSA  \cite{chelsa_bioclim_1,chelsa_bioclim_2}, Worldclim  \cite{worldclim_1,worldclim_2}, Soilgrids  \cite{soildgrid}, ASTER Global Digital Elevation Model  \cite{aster_elevation}, MODIS Terra+Aqua (land cover)  \cite{modis_land_cover}, DRYAD (human footprint)  \cite{dryad2016humanfootprint} and observations from GBIF and Pl@ntNet.

The time-series data (e.g., Landsat and CHELSA climatic variables) are also available as lightweight tensor \textit{cubes} where the information has been compressed down to geolocated points extraction based on species observation files. The easy loading of these cubes with PyTorch and their readiness make them a strong asset of MALPOLON's provided multi-modal datasets and baseline models.

\subsection{Use Cases}

While building MALPOLON, we have considered various scenarios and types of users. For the three most common "scenarios", we have prepared a set of experiments and comprehensive documentation. All three scenarios and how to work with them are further described below: \\

\noindent\textbf{"Custom train"}. The approach for custom dataset training is straightforward. Duplicate an experiment's main script, update its configuration file, and select a suitable data loader. If existing data loaders are insufficient, one could simply update the classes. Next, users choose a model by specifying the name and parameters in the config file (all TIMM models are available), and finalize the model parameters using custom or default values. Metrics, logs, and model weights are saved in a unique output folder. \\

\noindent\textbf{"Inference"}. For a scenario with just a pre-trained model inference, we made available all models available in TIMM or Torchvision. Besides, we provide selected foundational models such us GeoCLIP  \cite{vivanco2024geoclip}.
The recommended way to perform inference with a trained model on a test dataset is to use or duplicate an experiment’s main script and update the dataset path and the model’s weight path. Similarly to the “Custom train” scenario, users may import their modifications to data loaders and other classes before running the inference pipeline. Metrics, logs, and predictions are then outputted in a unique output folder. \\

\noindent\textbf{"Benchmarks"}. For a plug-and-play benchmarking on existing datasets such as those provided within the annual GeoLifeCLEF competition  \cite{geolifeclef2024} (organized on \href{https://www.kaggle.com/competitions/geolifeclef-2024}{Kaggle} in conjunction with \href{https://sites.google.com/view/fgvc11}{FGVC-CVPR} and \href{https://www.imageclef.org/LifeCLEF2024}{LifeCLEF} workshops \cite{joly2024lifeclef,lifeclef2024}), we provide data loaders and training examples to allow an easier start for anyone interested in deep species distribution modeling.
% the Institute of the National Institute for Research in Digital Science and Technology (INRIA). \\

\newpage
\subsection{Baseline Experiments and How to Use Them}
To make the framework accessible to ecologists, each experiment contains a CLI script ready to be run in the terminal with a single line command and a tutorial file detailing its purpose, how to configure it, the expected data, and running instructions. Three baseline experiments were provided for the "Custom train" and "Benchmark" scenarios for which users only need to update the following sections in the YAML file if needed:
\begin{itemize}
\itemsep0pt
\vspace{-0.1cm}
    \item \textbf{run}: parameters related to prediction and transfer learning
    \item \textbf{data}: data-related information, such as the path to a dataset or batch size
    \item \textbf{task}: type of experiment, e.g., binary, multiclass, or multilabel classification. 
    % The task determines the loss function used by the model: BCE loss for binary/multilabel classification and CE for multiclass classification.
    
    \item \textbf{trainer}: parameters used for tweaking a training or inference session 
    % and passed on to the PyTorch Lightning \verb|Trainer| class.
    
    \item \textbf{model}: defines the provider, architecture, and hyperparameters listed in the given provider's model builder. Additionally, it allows to call model modifiers to adapt their model structure. Namely, three modifiers are provided by MALPOLON to change a model's input and output shape.

    \item \textbf{optimizer}: defines the optimization algorithm, the loss and their hyperparameters
    % but it can be substituted in the configuration file.
    % \\ \verb|standard_prediction_systems.ClassificationSystem|).
\end{itemize}

\subsection{Assets}

The MALPOLON GitHub repository contains the mentioned datasets, models, baseline scripts and notebooks as well as instructions on how to install and use the framework for different use cases. Additionally, each "use case" contains detailed documentation explaining how to run and tune the experiment and giving insights about the data used by it. Additionally, there are handy data transformation and vizualization scripts being used to prepare the datasets used in many examples. Besides, we provide standardized code documentation, which is being updated every time new content is pushed to the main branch.

\section{Baseline performance}
\begin{figure}
    \centering
    {\hspace{4pt}}
    \includegraphics[width=0.9\linewidth]{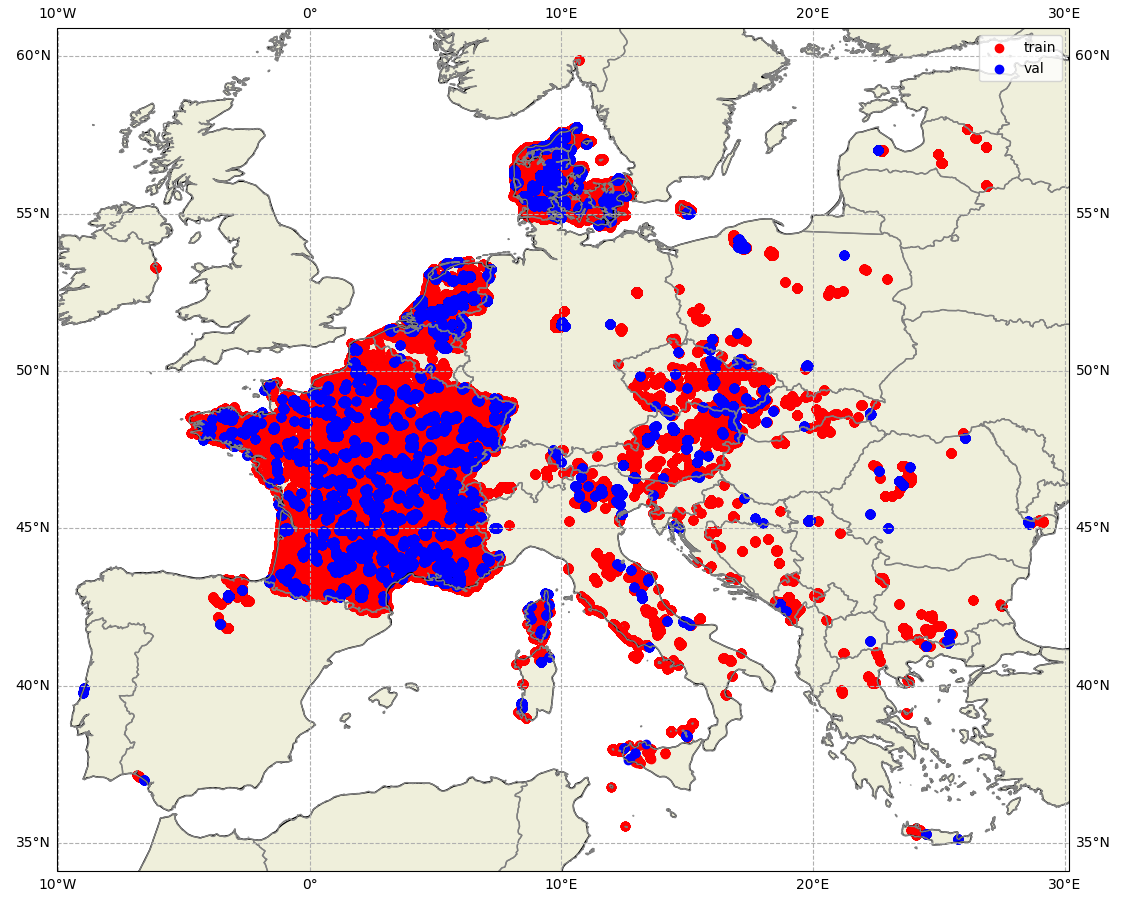}
    \caption{\textbf{Spatial split of training and validation data points.}}
    \label{fig:glc24_spatial_split}
\end{figure}

To test the suitability and versatility of MALPOLON, we opt to train on the data and models provided in the context of the recent GeoLifeCLEF competition  \cite{geolifeclef2024} organized in conjunction with \href{https://sites.google.com/view/fgvc11}{FGVC-CVPR} and \href{https://www.imageclef.org/LifeCLEF2024}{LifeCLEF} workshops \cite{joly2024lifeclef,lifeclef2024}. We have trained a multimodal ensemble model (MME) and single-modality models using the Sentinel-2A image patches, Landsat time series, and CHELSA bioclimatic time series \footnote{All architectures were provided by GeoLifeCLEF organizers.}.
The architecture consists of two ResNet-18  \cite{He2016Resnet} encoders for time series data and one Swin-v2 Transformer \cite{swin-v2} that encodes Sentinel-2A image patches; feature vectors are then concatenated and forwarded into a MLP.
For training, we used the GLC24 dataset \cite{picek2024geoplant,geolifeclef2024} consisting of 1.4M+ observations from 80k Presence Absence plots with 11k+ species scattered over Europe. The dataset is highly imbalanced as labels are characterized by a long-tailed distribution. The data have been spatially split into training (85\%) and testing (15\%) sets, using a spatial block holdout technique excluding training set points from validation "zones" of size 10 \textit{arcminutes}, which aims at reducing spatial bias. See Figure \ref{fig:glc24_spatial_split} for the geospatial distribution of the data. \\

\noindent\textbf{Experimental Settings:} To allow direct comparability, we use the same architectures as used in the provided GLC24 baselines, i.e., ResNet-18 for Landsat and Bioclimatic data and Swin-v2-b for Sentinel2 images. All ResNet and Swin models were fine-tuned from pre-trained ImageNet-1k weights. The ResNet-18 first convolutional layer input sizes were adapted to fit the expected size of the provided input variables, i.e., 3d cubes \footnote{Conv2D parameters, e.g., kernel size, stride, and padding, were also changed; for more details, refer to \cite{geolifeclef2024}}.
The MME model is a straightforward concatenation of the single-modality embeddings followed by two linear layers (first with a dropout with 10\% probability.). 

All models were trained for 20 epochs with a batch size of 64 using AdamW optimizer\footnote{\textit{AdamW} was parametrized with a learning rate of $2.5e10^{-4}$ and associated with a \textit{CosineAnnealingLR} scheduler parametrized with a max temperature of $25$}.
In order to predict multi-label species present in the plots, we used \textit{BCEWithLogitsLoss} with a sigmoid activation function. Additionally, in order to strengthen the rewarding process of correctly predicted species, the loss positive weight parameter was set to $10$.

Models are being evaluated by 2 main metrics adapted for imbalanced datasets: F1-score (F1) and ROC-AUC (as defined per Scikit-learn). The F1-score is more appropriate than a basic Accuracy as it puts more emphasis on False negatives and False positives. The ROC-AUC tracks the balance between True positives and False positives. Since we are most interested in avoiding False negatives, it makes sense to evaluate the model with such metrics. Additionally, we provide Precision (P) and Recall (R) to give more insight on the positive/negative rates. Finally, as each survey contains in average 20 species, we only compute F1, P and R on the Top-25 returned species.
\\

\noindent\textbf{Results}: As expected, a model trained using MALPOLON performed similarly to the model trained in the provided baseline notebook. Additionally, MALPOLON outputs the best model's weights, handles multi-process and multi-GPU usage seamlessly, and logs training progress to Tensorboard. Switching from training to prediction is as easy as changing two values in the config file. Finally, the framework provides convenient plotting functions to help visualize the model's predictions.
Regarding performances, since MALPOLON runs Python scripts, it is more efficient than notebooks, and PyTorch Lightning optimizes the distributed computations across CPU cores and GPU units. The main bottleneck resides in the datasets, which, when building custom ones, users should be mindful of making optimized. 

\begin{table}[h]
%\footnotesize
    \begin{center}
        \setlength{\tabcolsep}{0.35em} % for the horizontal padding
        \renewcommand{\arraystretch}{1.17}
        \begin{tabular}{@{}l cccc | cccc | cccc@{}}
        \hline
            % &\textbf{CHELSA + Landsat + Sentinel-2A}\\
             & \multicolumn{4}{c|}{\textit{Micro averaged}} & \multicolumn{4}{c|}{\textit{Sample averaged}} & \multicolumn{4}{c}{\textit{Macro averaged}} \\
            Model & AuC & P & R & F1 & AuC & P & R & F1 & AuC & P & R & F1 \\
            \hline
            ResNet-18$_l$ & 94.7 & \underline{23.4} & \underline{32.7} & \underline{27.3} & 94.8 & \underline{23.4} & \underline{39.1} & \underline{26.5} & 88.5 & 12.4 & 11.0 & \underline{7.7} \\
            \textit{ResNet-18$_b$} & \underline{95.1} & 21.1 & 29.4 & 24.5 & \underline{94.9} & 21.1 & 34.1 & 23.5 & \underline{90.0} & \underline{16.0} & \underline{12.1} & 7.6 \\
            \textit{Swin-v2$_s$} & 94.3 & 20.4 & 28.2 & 23.7 & 94.5 & 20.4 & 34.8 & 23.3 & 88.3 & 10.3 & 8.7 & 5.7 \\
            \textit{MME} & \textbf{96.7} & \textbf{26.2} & \textbf{36.5} & \textbf{30.5} & \textbf{96.3} & \textbf{26.2} & \textbf{43.4} & \textbf{29.6} & \textbf{93.2} & \textbf{18.9} & \textbf{14.5} & \textbf{9.4}  \\
        \hline
        \end{tabular}
    \end{center}
    \caption{\textbf{Baseline performance for selected custom architectures}. The Multimodal Ensemble (MME) model provides better performances than standard CNN and transformers on our metrics, i.e., Top-25 Precision (P), Top-25 Recall (R), Top-25 F1-score (F1), and Area under the Curve (AuC) computed in samples, macro and micro averaged. \textit{CNN$_l$}: ResNet-18 trained on Landsat data; \textit{CNN$_b$}: ResNet-18 trained on CHELSA bioclimatic data; \textit{Swin-v2$_s$}: Swin-v2 transformer trained on Sentinel-2A data.}
\label{table:results_baselines}
\vspace{-0.25cm}
\end{table}

\subsection{Benchmarking foundational models with MALPOLON}

Deep learning foundational models are trained on large amounts of broad data such that they can be applied across a wide range of use cases. As such, they are very much prized and beneficial to bootstrap topic-specific training as they allow reducing training times, which can range from hours to weeks if trained from scratch, depending on the data used. In the domain of species distribution modeling, such models are scarce, but interesting options, such as SatCLIP proposed by \citet{klemmer2023satclip} and GeoCLIP proposed by \citet{vivanco2024geoclip}, have emerged. Hereafter, we investigate the added value of these foundational models within the MME model in place of the Swin-v2 transformer, as well as an individual model to handle the Sentinel-2A modality. \\

\noindent\textbf{Results}: 
GeoCLIP outperforms the other two models in micro and sample-averaged metrics, with higher precision and recall, but like Swin-v2, it also falls short in macro-averaged performance, indicating limitations in its ability to generalize across all classes. On the other hand, SatCLIP underperformed heavily. 
The variant MME$^\dagger$, which replaces Swin-v2 with GeoCLIP, also demonstrates competitive performance, particularly excelling in precision metrics. It achieves the highest micro-averaged precision of 30.1\% and sample-averaged precision of 29.9\%, suggesting that integrating GeoCLIP enhances the model's ability to accurately identify positive instances. However, it shows slightly lower performance in macro-averaged metrics compared to the standard MME, indicating some variance in performance across different classes. For a more comprehensive evaluation, see Table \ref{table:results_foundational_models}.

\begin{table}[h]
%\footnotesize
    \begin{center}
        \setlength{\tabcolsep}{0.35em} % for the horizontal padding
        \renewcommand{\arraystretch}{1.17}
        \begin{tabular}{@{}l cccc | cccc | cccc@{}}
        \hline
            % &\textbf{CHELSA + Landsat + Sentinel-2A}\\
             & \multicolumn{4}{c|}{\textit{Micro averaged}} & \multicolumn{4}{c|}{\textit{Sample averaged}} & \multicolumn{4}{c}{\textit{Macro averaged}} \\
            Model & AuC & P & R & F1 & AuC & P & R & F1 & AuC & P & R & F1 \\
            \hline
            \textit{Swin-v2$_s$} & 94.3 & 20.4 & 28.2 & 23.7 & 94.5 & 20.4 & 34.8 & 23.3 & 88.3 & \underline{10.3} & \underline{8.7} & \underline{5.7} \\
            \textit{SatCLIP} & 89.5 & 11.5 & 14.5 & 12.8 & 89.5 & 10.4 & 13.3 & 11.1 & 85.0 & 0.9 & 2.4 & 1.0 \\
            \textit{GeoCLIP} & 95.1 & 24.1 & 30.4 & 26.9 & 95.1 & 22.8 & 30.1 & 24.6 & \underline{88.5} & 2.8 & 3.1 & 2.6 \\
            \hline
            \textit{MME} & \textbf{96.7} & \underline{26.2} & \underline{36.5} & \underline{30.5} & \textbf{96.3} & {26.2} & \textbf{43.4} & {29.6} & \textbf{93.2} & \textbf{18.9} & \textbf{14.5} & \textbf{9.4}  \\
            \textit{MME$^\dagger$} & \underline{95.8} & \textbf{30.1} & \textbf{37.9} & \textbf{33.5} & 96.0 & \textbf{29.9} & \underline{40.6} & \textbf{32.2} & \underline{88.5} & 3.6 & 3.9 & 3.4 \\
        \hline
        \end{tabular}
    \end{center}
    \caption{\textbf{MME baseline performance integrating foundational models} The MME model integrating GeoCLIP provides yet better performance in terms of micro and sample average evaluation, highlighting the relevance of foundational models paired with CNNs. The metrics used are the same as mentioned in Table \ref{table:results_baselines}. $^\dagger$MME with Swin-v2 replaced for GeoCLIP.}
\label{table:results_foundational_models}
\end{table}

\section{Conclusion}

In this work, we introduced "MALPOLON", a new PyTorch-based framework that enables ecologists to easily train deep-SDMs. The framework supports multi-GPU computation and provides access to neural networks to train models on classification and regression tasks to predict species distribution over geographical areas.
The framework is modular and follows encapsulation principles which makes it adapted for expert users knowledgeable about Python or PyTorch. However, it is also adapted for less experienced users coming from R, thanks to its online documentation, tutorial files and plug-and-play examples provided with the project's repository.
We showed that the framework can train baselines with complex deep architectures on a real-world use case, GeoLifeCLEF 2024, with strong metrics performance relative to other methods submitted on the \href{https://www.kaggle.com/competitions/geolifeclef-2024}{Kaggle challenge page}. Such architectures are then provided as part of the framework's available models and are coupled with additional examples for easy training.
Furthermore, the data used in all examples is made accessible for reproducibility and alleviates the need to manually gather and transform data from different sources.
The framework is open-sourced on \href{https://github.com/plantnet/malpolon/tree/main}{GitHub} and \href{https://pypi.org/project/malpolon/}{PyPi}, which enables it to use the wide Python community for future development.

\noindent\textbf{Limitations:}
As of now, too few models were trained using MALPOLON to provide a deep comparison analysis with the MME model. Likewise, the lack of similar open-sourced deep learning framework available makes it difficult to compare MALPOLON's features to those of existing frameworks, which do not operate using the same language or model types.

\noindent\textbf{Next Steps:}
Preliminary results have been conducted to predict \textit{EUNIS habitat types} instead of species. A convergence of efforts will be made to enable MALPOLON to support not only deep-SDM but also deep Habitat Distribution Modeling (deep-HDM).

\section*{Acknowledgement}
The research described in this paper was partly funded by the European Commission via the GUARDEN and MAMBO projects, which have received funding from the European Union’s Horizon Europe research and innovation program under grant agreements 101060693 and 101060639. The opinions expressed in this work are those of the authors and are not necessarily those of the GUARDEN or MAMBO partners or the European Commission.

\bibliography{bibliography}
\end{document}